\documentclass[10pt,conference,letterpaper]{IEEEtran}

\usepackage{times}  
\usepackage{helvet} 
\usepackage{courier}  
\usepackage{graphicx} 
\usepackage{subfigure}
\usepackage{amsmath,amssymb,amsfonts,bm}

\usepackage{algorithm}
\usepackage[noend]{algpseudocode}

\usepackage{url}
\usepackage[colorlinks,linkcolor=blue,anchorcolor=green,urlcolor=red,citecolor=red,pagebackref=true]{hyperref}

\usepackage{setspace}

\usepackage{multicol}
\usepackage{booktabs}
\usepackage{multirow}
\usepackage{threeparttable}
\usepackage{colortbl}

\begin{document}
\title{Federated Unlearning}

\author{\IEEEauthorblockN
{Gaoyang Liu$^1$,~~Xiaoqiang Ma$^1$,~~Yang Yang$^2$,~~Chen Wang$^1$,~~Jiangchuan Liu$^3$\\}
\IEEEauthorblockA{
\small $^1$Huazhong University of Science and Technology, Wuhan, China\\
$^2$Hubei University, Wuhan, China\\
$^3$Simon Fraser University, British Columbia, Canada\\
$^1$\{liugaoyang, maxiaoqiang, chenwang\}@hust.edu.cn,~$^2$yangyang@hubu.edu.cn,~$^3$jcliu@cs.sfu.ca
}}

\markboth{}
{G.~Liu \MakeLowercase{\textit{et al.}}: Federated Unlearning}

\maketitle

\begin{abstract}
Federated learning (FL) has recently emerged as a promising distributed machine learning (ML) paradigm. Practical needs of the ``right to be forgotten'' and countering data poisoning attacks call for efficient techniques that can remove, or unlearn, specific training data from the trained FL model. Existing unlearning techniques in the context of ML, however, are no longer in effect for FL, mainly due to the inherent distinction in the way how FL and ML learn from data. Therefore, how to enable efficient data removal from FL models remains largely under-explored. In this paper, we take the first step to fill this gap by presenting FedEraser\footnote{The code of FedEraser has been publicly released at\\ \url{https://www.dropbox.com/s/1lhx962axovbbom/FedEraser-Code.zip?dl=0}}, the first federated unlearning methodology that can eliminate the influence of a federated client's data on the global FL model while significantly reducing the time used for constructing the unlearned FL model. The basic idea of FedEraser is to trade the central server's storage for unlearned model's construction time, where FedEraser reconstructs the unlearned model by leveraging the historical parameter updates of federated clients that have been retained at the central server during the training process of FL. A novel calibration method is further developed to calibrate the retained updates, which are further used to promptly construct the unlearned model, yielding a significant speed-up to the reconstruction of the unlearned model while maintaining the model efficacy. Experiments on four realistic datasets demonstrate the effectiveness of FedEraser, with an expected speed-up of $\bm{4\times}$ compared with retraining from the scratch. We envision our work as an early step in FL towards compliance with legal and ethical criteria in a fair and transparent manner.
\end{abstract}




\section{Introduction}\label{Introduction}

\IEEEPARstart{A}{s} a distributed machine learning (ML) framework, federated learning (FL) has been recently proposed to address the problem of training ML models without direct access to diverse training data, especially for privacy-sensitive tasks~\cite{mcmahan2017communication,kone2018federated}.
FL allows multiple clients to jointly train a shared ML model by sending locally learned model parameter updates instead of their data to the central server.
With the distributed nature of such a computing paradigm, clients can thus benefit from obtaining a well-trained aggregated ML model while keeping their data in their own hands~\cite{yang2019federated,RN238,RN281}.

Given an FL model jointly trained by a group of clients, there are many settings where we would like
to remove specific training data from the trained model.
One example of the need is the ``right to be forgotten'' requirements enacted by recent legislations such as the General Data Protection Regulation (GDPR) in the European Union~\cite{RN245} and the California Consumer Privacy Act (CCPA) in the United States~\cite{harding2019ccpa}.
The ``right to be forgotten'' stipulates and sometimes legally enforces that individuals can request at any time to have their personal data cease to be used by a particular entity storing it.

Beyond the ``right to be forgotten'', data removal from FL models is also beneficial when certain training data becomes no longer invalid, which is especially of common occurrence due to the distributed learning methodology and inherently heterogeneous data distribution across parties.
Considering in FL some training data are polluted or manipulated by data poisoning attacks~\cite{xie2020dba,bagda2020backdoor,fung2020limitations}, or outdated
over time, or even identified to be mistakes after training.
The ability to completely forget such data and its lineage can greatly improve the security, responsiveness and reliability of the FL systems.

These practical needs call for efficient techniques that enable FL models to unlearn, or to forget what has been learned from the data to be removed (referred to as the \emph{target data}).
Directly delete the target data is proved to be unserviceable, as the trained FL model has potentially memorized the training data~\cite{RN222,melis2019exploiting,RN234}.
A naive way to satisfy the requested removal would be to simply retrain the model from scratch on the remaining data after removing the target one(s).
For many applications, however, the costs (in time, computation, energy, etc.) can be prohibitively expensive, especially when several rounds of alternations between training and aggregating among multiple participators are involved in FL settings.

Existing studies on unlearning in the context of ML (a.k.a. machine unlearning)~\cite{cao2015spmul,ginart2019making,bourtoule2020spmul,guo2020certified} cannot eliminate the influence of the target data on the global FL model either, mainly due to the inherent distinction in the way how FL and ML learn from data.
In particular, FL employs the \emph{iterative training} involving multiple rounds of training, where the clients' initial model for each round of training is obtained from the parameter updates in the previous round, and thus contains the information of all clients including the target one.
Such a \emph{forward coupling of information in parameter updates}
leads to fundamental challenges on the machine unlearning techniques, which are designed for the (arguably one round) centralized ML framework, yielding them no longer in effect for FL.
Therefore, how to efficiently forget the target data from FL models remains largely under-explored.

In this paper, we take the first step to fill this gap by presenting FedEraser, an efficient federated unlearning methodology that can eliminate the influences of a federated client's data on the global model while significantly reducing the unlearning time.
The basic idea of FedEraser is to trade the central server's storage for unlearned model's construction time, where FedEraser reconstructs the unlearned model by leveraging the historical parameter updates of clients that have been retained at the central server during the training process of FL.
Since the retained updates are derived from the global model which contains the influence of the target client's data, these updates have to be calibrated for information decoupling before using them for unlearning. Motivated by the fact that the client updates indicate in which direction the parameters of the global model need to be changed to fit the model to the training data~\cite{nasr2019comprehensive}, we further calibrate the retained client updates through performing only a few rounds of calibration training to approximate the direction of the updates without the target client, and the unlearned model can be constructed promptly using the calibrated updates.

We summarize our major contributions as follows:
\begin{itemize}
 \item We frame the problem of federated unlearning, and present FedEraser, the first efficient unlearning algorithm in FL that enables the global model to ``forget" the target client's data. FedEraser is non-intrusive and can serve as an opt-in component inside existing FL systems.

 \item We develop novel storage-and-calibration techniques to tackle the forward coupling of information in parameter updates, which can provide a significant speed-up to the reconstruction of the unlearned model while maintaining the model efficacy.

 \item We propose a new indicator by using the layer parameter's deviation between the unlearned model and the retrained model, to measure the effectiveness of FedEraser on the global model.

 \item We evaluate the performance of FedEraser on four realistic datasets, and compare it with two baselines. The results demonstrate that FedEraser can remove the influence of the target client's data, with an expected speed-up of $4\times$ compared with retraining from the scratch.

\end{itemize}


\section{Preliminary} \label{Sec:Preliminary}

\subsection{Federated Learning}
FL is proposed recently by Google~\cite{konevcny2016federated1,konevcny2016federated2} as a promising solution that can train a unified deep learning (DL) model across multiple decentralized clients holding local data samples, under the coordination of a central server. In FL, each client's data is stored on its local storage and not transferred to other clients or the central server; only locally learned model parameter updates are exposed to the central server for aggregation to construct the unified global model.
FL embodies the principles of focused collection and data minimization, and can mitigate many of the systemic privacy risks resulting from traditional, centralized DL approaches.


Although there are different forms of FL, most existing works mainly focus on the horizontal FL, or sample-based FL, in which case datasets of multiple clients share the same feature space but different space in samples. On the contrary, vertical FL, or feature-based FL, is applicable to the scenarios in which datasets coming from different domains share the same sample space but differ in feature space.
We focus on horizontal FL in this paper.

\subsection{Architecture of FL}
In a typical architecture of FL, there are $K$ federated clients with the same data structure and feature space that collaboratively train a unified DL model with the coordination of a central server. The central server organizes the model training process, by repeating the following steps until the training is stopped. At the $i_{th}$ training round ($i \in E$, where $E$ is the number of training rounds), the updating process of the global model $\mathcal{M}$ is performed as follows:

\textbf{Step 1. } Federated clients download the current global model $\mathcal{M}^i$ and the training setting from the central server.

\textbf{Step 2. } Each client $C_k\ (k \in \{1,2,\cdots, K\})$ trains the downloaded model $\mathcal{M}^i$ on its local data $D_k$ for $E_{local}$ rounds based on the training setting, and then computes an update $U_k^i$ with respect to $\mathcal{M}^i$.

\textbf{Step 3. } The central server collects all the updates $U^i=\{U_1^i,U_2^i,\cdots,U_K^i\}$ from the $K$ clients.

\textbf{Step 4. } The central server updates the global model on the basis of the aggregation of the collected updates $U^i$, thereby obtaining an updated model $\mathcal{M}^{i+1}$ that will play the role as the global model for the next training round.

When the termination criterion has been satisfied, the central server will stop the above iterative training process and get the final FL model $\mathcal{M}$.

\section{Design of FedEraser}\label{Sec:Design}

\subsection{Overview}

In order to efficiently eliminate the influence of the target client's data from the trained global model, we add one extra function to the central server for FedEraser in the current architecture of FL, while the original functions of the central server remaining unchanged.
Specifically, during the training process of the global model, the central server retains the updates of the clients, at intervals of regular rounds, as well as the index of corresponding round, so as to further calibrate the retained updates to reconstruct the unlearned global model, rather than retraining from scratch.

For clarity, we denote the round intervals as $\Delta t$, and the retained updates of the client $C_k$ as $U_k^{t_j}$ ($j \in \{1,2,\cdots, T\}$, where $t_1=1$, $t_{j+1}=t_j+\Delta t$, $T$ is the number of retaining rounds that equals to $\lfloor \frac{E}{\Delta t} \rfloor$, and $\lfloor \cdot \rfloor$ is the floor function). Thus, the whole retained updates can be denoted as $U = \{U^{t_1}, U^{t_2}, \cdots, U^{t_T}\}$, where $U^{t_j} = \{U_1^{t_j}, U_2^{t_j}, \cdots U_K^{t_j}\}$.

Given the retained updates $U$ and the target client $C_{k_u}$ whose data is required to be removed from the FL model, FedEraser mainly involves the following four steps:
(1) \emph{calibration training}, (2) \emph{update calibrating}, (3) \emph{calibrated update aggregating}, and (4) \emph{unlearned model updating}. The first step is performed on the calibrating clients $C_{k_c}$ ($k_c\in[1,2,\cdots,K]\setminus k_u$, i.e., the federated clients excluding the target one), while the rest steps are executed on the central server (c.f.~Algorithm~\ref{algorithm1}).

\addtolength{\topmargin}{0.05in}
\subsection{Design Details}
\subsubsection{Calibration Training}

Specifically, at the $t_j$th training round, we let the calibrating clients run $E_{cali}$ rounds of local training with respect to the calibrated global model $\widetilde{\mathcal{M}}^{t_j}$ that is obtained by FedEraser in the previous calibration round.

It should be noticed that FedEraser can directly update the global model without calibration of the remaining clients' parameters at the first reconstruction epoch. The reason for this operation is that the initial model of the standard FL has not been trained by the target client and thus this model does not contain the influence brought by the target client.

After the calibration training, each calibrating client $C_{k_c}$ calculates the current update $\widehat{U}_{k_c}^{t_j}$ and sends it to the central server for update calibrating.

\subsubsection{Update Calibrating}
After the calibration training, the central server can get each client's current update $\widehat{U}_{k_c}^{t_j}$ with respect to the calibrated global model $\widehat{\mathcal{M}}^{t_j}$. Then FedEraser leverages $\widehat{U}_{k_c}^{t_j}$ to calibrate the retained update $U_{k_c}^{t_j}$.
In FedEraser, the norm of $U_{k_c}^t$ indicates how much the parameters of the global model needs to be changed, while the normalized $\widehat{U}_{k_c}^{t_j}$ indicates in which direction the parameters of $\widetilde{\mathcal{M}}^{t_j}$ should be updated.
Therefore, the calibration of $U_{k_c}^{t_j}$ can be simply expressed as:
\begin{equation}
	\label{equ:1}
		\widetilde{U}_{k_c}^{t_j} = |U_{k_c}^{t_j}| \frac{\widehat{U}_{k_c}^{t_j}}{||\widehat{U}_{k_c}^{t_j}||}
\end{equation}

\subsubsection{Calibrated Update Aggregating}
Given the calibrated client updates $\widetilde{U}^{t_j}=\{\widetilde{U}_{k_c}^{t_j}|k_c\in[1,2,\cdots,K]\setminus k_u\}$, FedEraser next aggregates these updates for unlearned model updating.
In particular, FedEraser directly calculates the weighted average of the calibrated updates as follows:
\begin{equation}
	\label{equ:2}
	\begin{aligned}
		\widetilde{\mathcal{U}}^{t_j} = \frac{1}{(K-1)\sum_{k_c} w_{k_c}} \sum_{k_c} w_{k_c} \widetilde{U}_{k_c}^{t_j}
	\end{aligned}
\end{equation}
where $w_{k_c}$ is the weight for the calibrating client obtained from the standard architecture of FL, and $w_{k_c} = \frac{N_{k_c}}{\sum_{k_c} N_{k_c}}$ where $N_{k_c}$ is the number of records the client $C_{k_c}$ has.
It is worth noting that this aggregation operation is consistent with the standard FL.

\begin{algorithm}[t]
    \caption{FedEraser}
	\label{algorithm1}
    {\fontsize{10}{12}\selectfont
	\begin{algorithmic}
		\Require{Initial global model $\mathcal{M}^1$; retained client updates $U$}
		\Require{Target client index $k_u$}
		\Require{Number of global calibration round $T$}
		\Require{Number of local calibration training epoch $E_{cali}$}
		
		\noindent \textbf{{{Central server executes}:}}
		\For{each round $R_{t_j}, j\in \{1,2,\cdots, T\}$} 
		
		\For{ each client $C_{k_c}, k_c\in \{1,2, \cdots, K\}\setminus k_u $ \textbf{in parallel}}
		\State{$\widehat{U}_{k_c}^{t_j}$ $\leftarrow$ CaliTrain($C_{k_c},\widetilde{\mathcal{M}}_{k_c}^{t_j},E_{cali}$)}
		\State{$\widetilde{U}_{k_c}^{t_j}$ $\leftarrow$ $|U_{k_c}^{t_j}|  \frac{\widehat{U}_{k_c}^{t_j}}{||\widehat{U}_{k_c}^{t_j}||}$} \Comment{Update Calibrating}
		\EndFor	
		\State{\textbf{end}}	
		
		\State{$\widetilde{\mathcal{U}}^{t_j}$ $\leftarrow$ $ \frac{1}{(K-1)\sum w_{k_c}} \sum_{{k_c}} w_{k_c} \widetilde{U}_{k_c}^{t_j}$}\Comment{Update Aggregating}
		
		\State{$\widetilde{\mathcal{M}}^{t_{j+1}} \leftarrow  \widetilde{\mathcal{M}}^{t_j} + \widetilde{\mathcal{U}}^{t_j}$}\Comment{Model Updating}	
		\EndFor
		\State{\textbf{end}}	
		
		\State
		
		\noindent \textbf{CaliTrain($C_{k_c},\widetilde{\mathcal{M}}_{k_c}^{t_j},E_{cali}$)}:  \quad//~Run on client $C_{k_c}$
		\For{each local training round $j$ from $1$ to $E_{cali}$}
		\State{$\widetilde{\mathcal{M}}_{k_c}^{t_j}|_{j+1} \leftarrow Train(\widetilde{\mathcal{M}}_{k_c}^{t_j}|_j, D_{k_c})$}
		\EndFor
		\State{\textbf{end}}
		
		\State{$\widehat{U}_{k_c}^{t_j}$ $\leftarrow$ Calculating Update($\widetilde{\mathcal{M}}_{k_c}^{t_j}|_{E_{cali}}$, $\widetilde{\mathcal{M}}_{k_c}^{t_j}|_1$)}
		\State{\textbf{return} $\widehat{U}_{k_c}^{t_j}$ to the central server}
\end{algorithmic}
}
\end{algorithm}

\subsubsection{Unlearned Model Updating}
With the aggregation of the calibrated updates, FedEraser can thus renovate the global FL model as:
\begin{equation}
	\label{equ:3}
	\begin{aligned}
		 \widetilde{\mathcal{M}}^{t_{j+1}} = \widetilde{\mathcal{M}}^{t_j} + \widetilde{\mathcal{U}}^{t_j}
	\end{aligned}
\end{equation}
where $\widetilde{\mathcal{M}}^{t_j}$ (resp. $\widetilde{\mathcal{M}}^{t_{j+1}}$) is the current global model (resp. updated global model) calibrated by FedEraser.

The central server and the calibrating clients collaboratively repeat the above process, until the original updates $U$ have all been calibrated and then updated to the global model $\widetilde{\mathcal{M}}$.
Finally, FedEraser gets the unlearned global model $\widetilde{\mathcal{M}}$ that has removed the influence of the client $C_{k_u}$'s data.

Once the unlearned global model $\widetilde{\mathcal{M}}$ is obtained, the standard deployment process of the deep learning model can be performed, including manual quality assurance, live A/B testing (by using the unlearned model $\widetilde{\mathcal{M}}$ on some clients' data and the original model $\mathcal{M}$ on other clients' data to compare their performance).

It is worth noting that FedEraser does not require far-reaching modifications of neither the existing architecture of FL nor the training process on the federated clients, yielding it very easy to be deployed in existing FL systems.
In particular, the process of calibration training executed on the federated clients can directly reuse the corresponding training process in the standard FL framework.
The aggregating and updating operations in FedEraser do not need to modify the existing architecture of FL, while only the additional retaining functionality is required at the central server side.
In addition, FedEraser can be performed unwittingly, as it does not involve any information about the target client, including his/her updates and local data, during the unlearning process.

\subsection{Time Consumption Analysis}
One crucial feature of FedEraser is that it can speed up the reconstruction of the unlearned model, compared with retraining the model from scratch. Thus, we provide an elementary analysis of the speed-up significance of FedEraser here.
For ease of presentation, we use the time consumption required for retraining from scratch as the baseline.

In FedEraser, there are two settings that can speed up the reconstruction of the unlearned model.
First, we modify the standard FL to retain the client updates at intervals of regular rounds. We use a hyper-parameter $\Delta t$ to control the size of the retaining interval. Since FedEraser only processes on retained updates, the larger $\Delta t$ is, the less retaining rounds are involved, and the less reconstruction time FedEraser would require. This setting could provide FedEraser with a speed-up of $\Delta t$ times.
Second, FedEraser only requires the calibrating client perform a few rounds of local training in order to calibrate the retained updates. Specifically, the round number of the calibration training is controlled by the calibration ratio $\verb"r" = E_{cali}/E_{loc}$. This setting can directly reduce the time consumed by training on the client, and provide FedEraser with a speed-up of $\verb"r"^{-1}$ times.
Overall, FedEraser can reduce the time consumption by $\verb"r"^{-1}\Delta t$ times compared with retraining from scratch.

In our experiments, we empirically find that when $r=0.5$ and $\Delta t=2$, FedEraser can achieve a trade-off between the performance of the unlearned model and the time consumption of model reconstruction (detailed in the following section). In such a case, FedEraser can achieve an expected speed-up of $4 \times $ compared with retraining from the scratch.

\section{Performance Evaluation}\label{Sec:Evaluation}
In this section, we evaluate the performance of FedEraser on different datasets and models. Besides, we launch membership inference attacks (MIAs) against FedEraser to verify its unlearning effectiveness from a privacy perspective.

\subsection{Experimental Setup}
\subsubsection{Datasets} We utilize four datasets in our experiments, including UCI
Adult\footnote{\url{https://archive.ics.uci.edu/ml/datasets/Adult}}, Purchase\footnote{\url{https://www.kaggle.com/c/acquire-valued-shoppers-challenge/data}}, MNIST\footnote{\url{http://yann.lecun.com/exdb/mnist}}, and CIFAR-10\footnote{\url{http://www.cs.toronto.edu/~kriz/cifar.html}}.

\textbf{\emph{Adult (Census Income). }}This dataset includes $48,842$ records with $14$ attributes such as age, gender, education, marital status, occupation, working hours, and native country. The classification task of this dataset is to predict if a person earns over $\$50K$ a year based on the census attributes.

\textbf{\emph{Purchase. }}Purchase dataset is obtained from Kaggle's ``acquire valued shoppers" challenge whose purpose is to design accurate coupon promotion strategies.  Purchase dataset contains shopping histories of several thousand shoppers over one year, including many fields such as product name, store chain, quantity, and date of purchase.
In particular, Purchase dataset (with $197,324$ records) does not contain any class labels. Following~\cite{bourtoule2020spmul,shokri2017membership,salem2018ml}, we adopt an unsupervised clustering algorithm to assign each data record with a class label. We cluster the records in Purchase dataset into 2 classes.

\textbf{\emph{MNIST. }}This is a dataset of $70,000$ handwritten digits formatted as $32 \times 32$ images and normalized so that the digits are located at the central of the image. It includes sample images of handwritten digits from $0$ to $9$. Each pixel within the image is represented by 0 or 1.

\textbf{\emph{CIFAR-10. }}CIFAR-10 is a benchmark dataset used to evaluate image recognition algorithms. This dataset consists of $60,000$ color images of size $32\times32$ and has $10$ classes such as ``air plane", ``dogs", ``cats", and etc. Particularly, CIFAR-10 is a balanced dataset with $6,000$ randomly selected images for each class. Within CIFAR-10 dataset, there are $50,000$ training images and $10,000$ testing images.

\subsubsection{Global Models} In the paradigm of FL and Federated Unlearning, the global model will be broadcasted to all clients and serve as the initial model for each client's training process. We make use of 4 global models with different structures for different classification tasks. The details of these models are summarized in Table~\ref{tab_model_structure}, where FC layer means fully connected layer in the deep neural network (DNN) models, and Conv. (resp. Pool.) layer represents convolutional (resp. maxpooling) layer in the convolutional neural network (CNN) models.

\renewcommand\arraystretch{1.2}
\begin{table}[t]
	\caption{The architectures of federated models. }
	\centering
    {\fontsize{9.5}{12}\selectfont
    \setlength{\tabcolsep}{5mm}{
	\begin{tabular}{!{\vrule width 1.0pt}l|c!{\vrule width 1.0pt}}
	\specialrule{1.0pt}{0pt}{0pt}
	\textbf{Dataset}	&  \textbf{Model Architecture}  \\ \specialrule{0.6pt}{0pt}{0pt}
	Adult	&  2 FC layers \\ \specialrule{0.6pt}{0pt}{0pt}
	Purchase 	&  3 FC layers\\ \specialrule{0.6pt}{0pt}{0pt}
	MNIST	& 2 Conv. and 2 FC layers \\ \specialrule{0.6pt}{0pt}{0pt}
	CIFAR-10	& 2 Conv., 2 Pool., and 2 FC layers $  $  \\ \specialrule{1.0pt}{0pt}{0pt}
	\end{tabular}
    }}
\label{tab_model_structure}
\end{table}

\begin{figure*}[t]
	\centering
	\subfigure[Prediction Accuracy (Testing Data)]{
		\includegraphics[width=7.7cm]{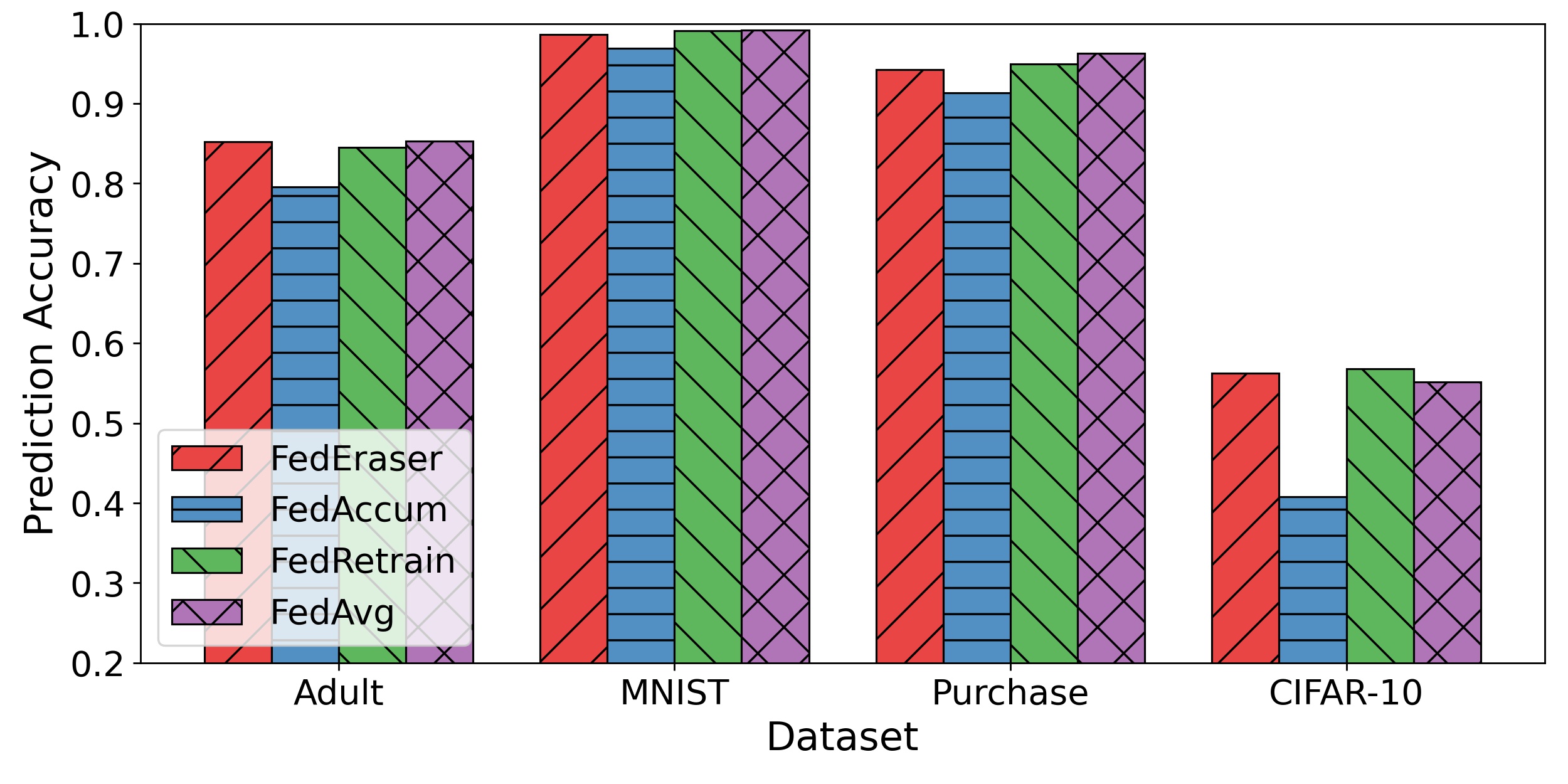}
		\centering
		\label{Figure:acc_on_test}
	}
	\subfigure[Prediction Accuracy (Target Data)]{
		\includegraphics[width=7.7cm]{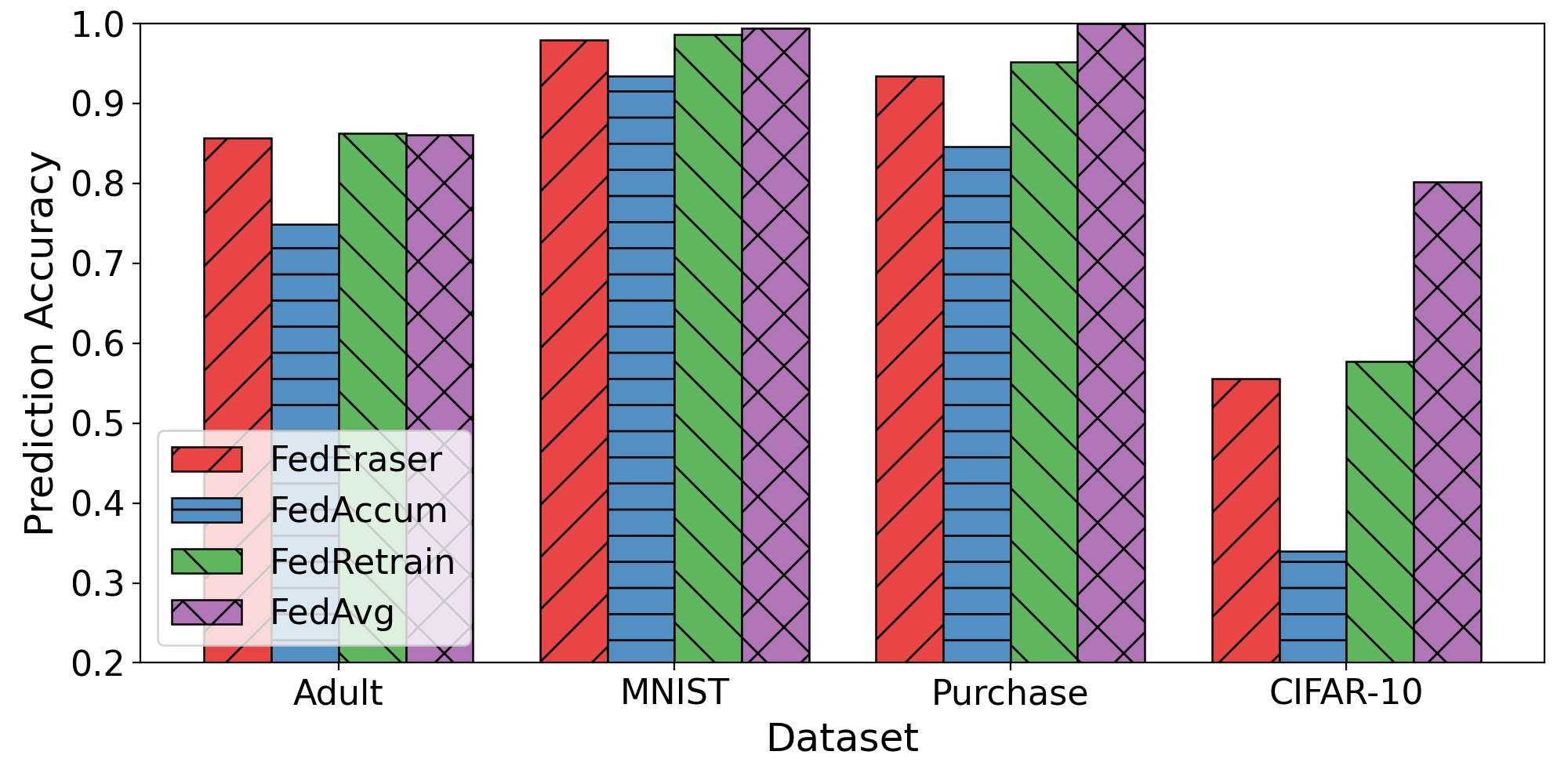}
		\centering
		\label{Figure:acc_on_client}
	}
	\caption{Prediction performance comparison.}
	\label{Figure:acc}
\end{figure*}

\subsubsection{Evaluation Metrics} We evaluate the performance of FedEraser using standard metrics in the ML field, including the \textbf{\emph{accuracy}} and the \textbf{\emph{loss}}. We also measure the unlearning \textbf{\emph{time}} consumed by FedEraser to make a given global model forget one of the clients.

Furthermore, in order to assess whether or not the unlearned model still contains the information about the target client, we adopt the following three extra metric.
One metric is the \textbf{\emph{prediction difference}}, denoted as the L2 norm of prediction probability difference, between the original global model and the unlearned model:
\begin{equation}
	P_{diss} = \frac{1}{N} \sum_{i=1}^{N} ||\mathcal{M}(x_i) - \widetilde{\mathcal{M}}(x_i)||_2~~~~x_i \in D_{k_u} 
\end{equation}
where $N$ is the number of the target client's samples $D_{k_u}$. $\mathcal{M}(x_i)$ (resp. $\widetilde{\mathcal{M}}(x_i)$) is the prediction probability of the sample $x_i$ obtained from the original (resp. unlearned) model.

The rest two metrics are obtained from the MIAs that we perform against the unlearned global model. The goal of MIAs is to determine whether a given data was used to train a given ML model. Therefore, the performance of MIAs can measure the information that still remains in the unlearned global model.
We utilize the \textbf{\emph{attack precision}} of the MIAs against the target data as one metric, which presents the proportion of target client's data that are predicted to have been participated in the training of the global model.
We also use the \textbf{\emph{attack recall}} of the MIAs, which presents the fraction of the data of the target client that we can correctly infer as a part of the training dataset. In other words, attack precision and attack recall measure the privacy leakage level of the target client.

\subsubsection{Comparison Methods} In our experiments, we compare FedEraser with two different methods: Federated Retrain (FedRetrain) and Federated Accumulating (FedAccum).

\textbf{FedRetrain:}
a simple method for unlearning by retraining this model from scratch on the remaining data after removing the target client's data, which will serve as the \textbf{baseline} in our experiments. Empirically, FedRetrain provides an \textbf{upper bound} on the prediction performance of the unlearned model reconstructed by FedEraser.

\textbf{FedAccum:}
a simple method for unlearning by directly accumulating the previous retained updates of the calibrating clients' local model parameters, and leveraging the accumulated updates to update the global model.
The update process can be expressed as follows:
\begin{equation}
	\widetilde{M}_{accum}^{t_{j+1}} = \widetilde{M}_{accum}^{t_j} +  \widetilde{\mathcal{U}}_{accum}^{t_j}
\end{equation}
where $\widetilde{\mathcal{U}}_{accum}^{t_j}$ is the accumulation of the model updates at $t_j$th round, and $\widetilde{\mathcal{U}}_{accum}^{t_j}=\frac{1}{(K-1)\sum w_{k_c}} \sum_{{k_c}} w_{k_c} U_{k_c}^{t_j}$.
Besides, $M_{accum}^{t_j}$ (resp. $M_{accum}^{t_{j+1}}$) represents the global model before (resp. after) updating with $\widetilde{\mathcal{U}}_{accum}^{t_j}$.
The main difference between FedEraser and FedAccum is that the latter does not calibrate the clients' updates.

In order to evaluate the utility of the unlearned global model, we also compare FedEraser with the classical FL without unlearning. We employ the most widely used FL algorithm, federated averaging (\textbf{FedAvg})~\cite{mcmahan2017communication}, to construct the global model.
FedAvg executes the training  procedure in parallel on all federated clients and then exchanges the updated model weights. The updated weights obtained from every client are averaged to update the global model.

\subsubsection{Experiment Environment}
In our experiments, we use a workstation equipped with an Intel Core i7 9400 CPU and NVIDIA GeForce GTX 2070 GPU for training the deep learning models. We use Pytorch 1.4.0 as the deep learning framework with CUDA 10.1 and Python 3.7.3.

We set the number of clients to $20$, the calibration ratio $\verb"r" = E_{cali}/E_{loc}=0.5$, and the retaining interval $\Delta t=2$. As for other training hyper-parameters, such as learning rate, training epochs, and batch size, we use the same settings to execute our algorithm and the comparison methods.

\renewcommand\arraystretch{1.2}
\begin{table*}[t]
	\caption{Time consumption of federated model construction.}
	\centering
	\label{Tab:time}
	{\fontsize{9.5}{12}\selectfont
    \setlength{\tabcolsep}{5mm}{
    \begin{tabular}{!{\vrule width 1.0pt}l||c|c|c|c!{\vrule width 1.0pt}}
        \specialrule{1.0pt}{0pt}{0pt}
        \multirow{2}{*}{\textbf{Method}} & \multicolumn{4}{c!{\vrule width 1.0pt}}{\textbf{Time Consumption (second)}} \\ \cline{2-5}
		&Adult     &MNIST     &Purchase     &CIFAR-10     \\ \hline
		FedAvg		       &$1.74\times 10^{2}$      &$1.16\times 10^{3}$      &$3.87\times 10^{2}$     &$3.38\times 10^{3}$    \\ \hline
		FedRetrain 		        &$1.69\times 10^{2}$      &$1.08\times 10^{3}$      &$3.75\times 10^{2}$     &$3.01\times 10^{3}$     \\ \hline
		FedAccum		&$2.14\times 10^{-2}$     &$6.2\times 10^{-1}$      &$1.56\times 10^{-1}$     &$1.29\times 10^{-1}$     \\ \hline
		\textbf{FedEraser}		 &\textbf{$4.97\times 10^{1}$}      &\textbf{$2.23\times 10^{2}$}      &\textbf{$9.21\times 10^{1}$}     &\textbf{$7.51\times 10^{2}$}      \\
        \specialrule{1.0pt}{0pt}{0pt}
	\end{tabular}
    }}
\end{table*}

\renewcommand\arraystretch{1.2}
\begin{table*}[t]
	\caption{Prediction loss on the target client's data.}
	\label{Tab:loss}
	\centering
    {\fontsize{9.5}{12}\selectfont
    \setlength{\tabcolsep}{5mm}{	
	\begin{tabular}{!{\vrule width 1.0pt}l||c|c|c|c!{\vrule width 1.0pt}}
		 		\specialrule{1.0pt}{0pt}{0pt}
	 		\multirow{2}{*}{\textbf{Method}} & \multicolumn{4}{c!{\vrule width 1.0pt}}{\textbf{Prediction Loss of Target Data}} \\ \cline{2-5}
						&Adult     &MNIST     &Purchase     &CIFAR-10     \\ \hline
	 		FedAvg		       &$5.26\times 10^{-3}$      &$3.85\times 10^{-4}$      &$2.49\times 10^{-5}$     &$1.19\times 10^{-2}$    \\ \hline
			FedRetrain  		        &$5.49\times 10^{-3}$      &$7.85\times 10^{-4}$      &$2.28\times 10^{-3}$     &$3.51\times 10^{-2}$     \\ \hline
		 		FedAccum		&$1.45\times 10^{-2}$     &$1.26\times 10^{-3}$      &$1.45\times 10^{-2}$     &$1.66\times 10^{-1}$     \\ \hline
		 		\textbf{FedEraser}		 &\textbf{$5.42\times 10^{-3}$}      &\textbf{$1.03\times 10^{-3}$}      &\textbf{$3.85\times 10^{-3}$}     &\textbf{$2.03\times 10^{-2}$}      \\ \specialrule{1.0pt}{0pt}{0pt}
		 	\end{tabular}
    }}
	\end{table*}

\subsection{Performance of FedEraser}\label{Exp.1}
In this section, we evaluate the performance of FedEraser from two perspectives: model utility and client privacy. We have to emphasize here that there is no overlap between the testing data and the target data.

\subsubsection{Performance on Testing Data}
Fig.~\ref{Figure:acc_on_test} shows the prediction accuracy of FedEraser and three comparisons on the testing data and the target data.
From the results we can see that FedEraser performs closely as FedRetrain (baseline) on all datasets, with an average difference of only $0.61\%$.
Especially, for Adult dataset, FedEraser achieves a prediction accuracy of $0.853$, which is higher than that of FedAccum by $5.76\%$ and lower than that of FedRetrain by $0.8\%$. On MNIST dataset, the performance of FedEraser achieves an accuracy of
$0.986$, which only has $0.52\%$ difference from that of FedRetrain. For Purchase dataset, FedEraser can achieve a testing accuracy of $0.943$. FedAccum and FedRetrain achieve a mean testing accuracy of $0.913$ and $0.949$, respectively. As for CIFAR-10 dataset, FedEraser gets a testing accuracy of $0.562$, which is lower than that of FedRetrain by $ 0.52\%$. In such a case,  FedAccum only achieves a testing accuracy of $0.408$. Overall, FedEraser can achieve a prediction performance close to FedAvg and FedRetrain, but better than that of FedAccum, indicating high utility of the obtained unlearned model.

Table~\ref{Tab:time} shows the time consumption of FedEraser and the comparison methods in constructing the global models. According to the results, it is obvious that FedRetrain takes the same order of magnitude of the time as FedAvg to reconstruct the global model. On the contrary, FedEraser can significantly speed up the removal  procedure of the global model, improving the time consumption by $3.4\times$ for Adult dataset. As for MNIST and Purchase datasets, FedEraser reduces the reconstruction time by $4.8\times$ and $4.1\times$, respectively. Besides, FedEraser also provides a speed-up of $4.0\times$ in reconstructing the global model for complex classification tasks.
As for FedAccum, since it only aggravates the retained parameters of the calibrating client's models in every global epoch and updates the global model with the aggravations, it does not involve the training process on the clients. Consequently,  FedAccum could significantly reduce the time consumption of model reconstruction, but at the cost of the prediction accuracy.

\subsubsection{Performance on Target Data}
In addition, we compare the prediction performance of FedEraser and the comparison methods on the target client's data. The experiment results are shown in Fig.~\ref{Figure:acc_on_client}.
For the target data, FedEraser achieve a mean prediction accuracy of $0.831$ over all datasets, which is close to that of FedRetrain but much lower than that of FedAvg. Compared with FedAccum, FedEraser performs $11.5\%$ better. As shown in Fig.~\ref{Figure:acc_on_client}, on the Adult and MNIST datasets, the performance of our method is slightly worse than baseline by $0.52\%$ and $0.65\%$ respectively. However, in these two cases, FedEraser still performs better than FedAccum by $10.9\%$ and $4.54\%$. For Purchase dataset, FedEraser can achieve a mean accuracy of $0.934$ on the target client's data, which is better than that of FedAccum by $8.82\%$. Nevertheless, FedRetrain achieves an accuracy of $0.952$ on the target data and performs better than FedEraser by $1.8\%$. As for the performance on the target client's data of CIFAR-10, FedEraser obtains a mean accuracy of $0.556$ while FedRetrain achieves $0.577$. However, FedAccum only can get a prediction accuracy of $0.339$ on the target data. In general, an ML model has a higher prediction accuracy of  the training data than that of testing data. Therefore, the prediction similarity between the unlearned model and the retrained model further reflects the removal effectiveness of FedEraser.

\begin{figure*}[t]
	\centering
	\subfigure[Attack Precision]{
		\includegraphics[width=7.7cm]{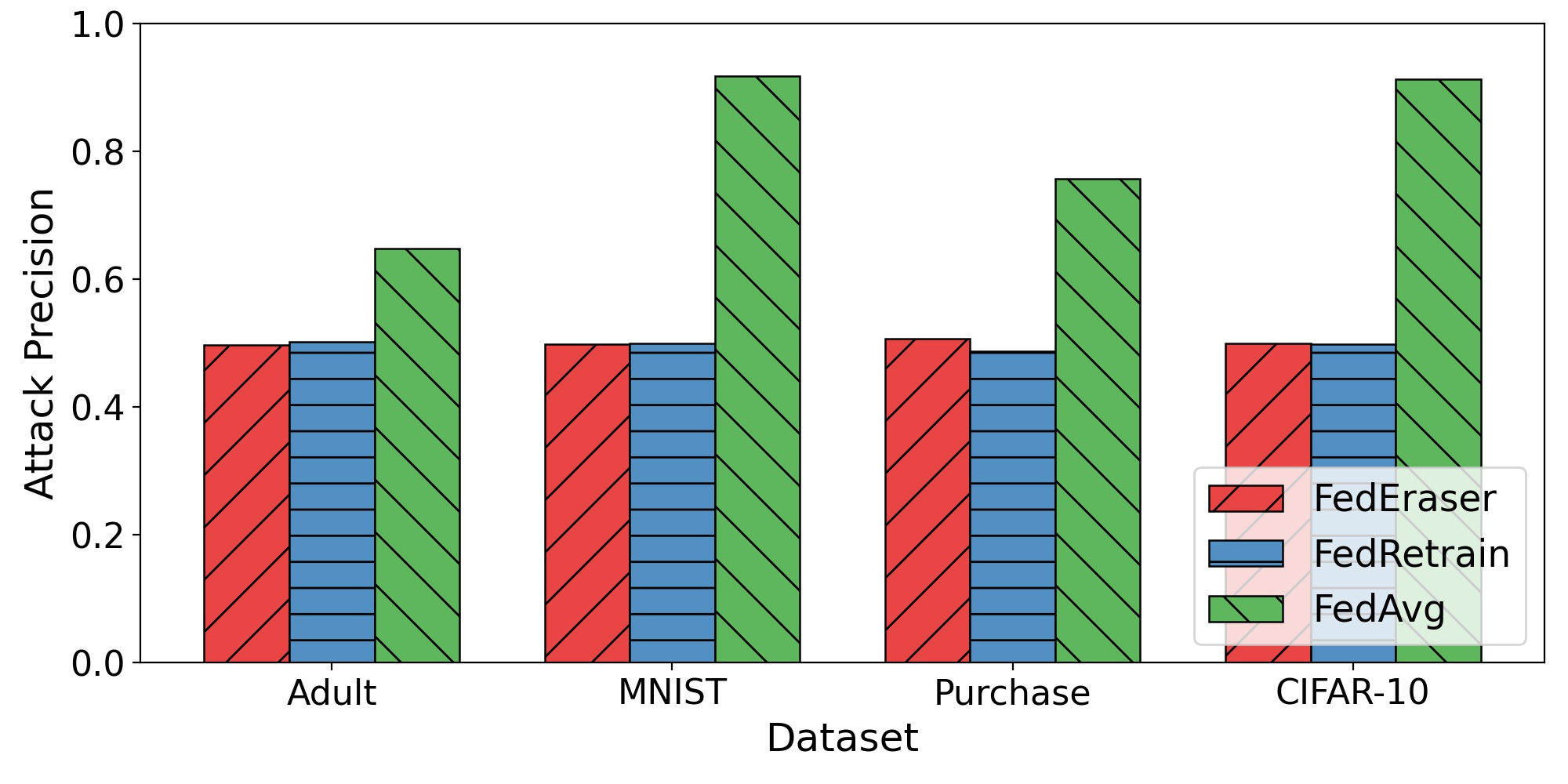}
		\centering
		\label{Figure:mia_acc}
	}
	\subfigure[Attack Recall]{
		\includegraphics[width=7.7cm]{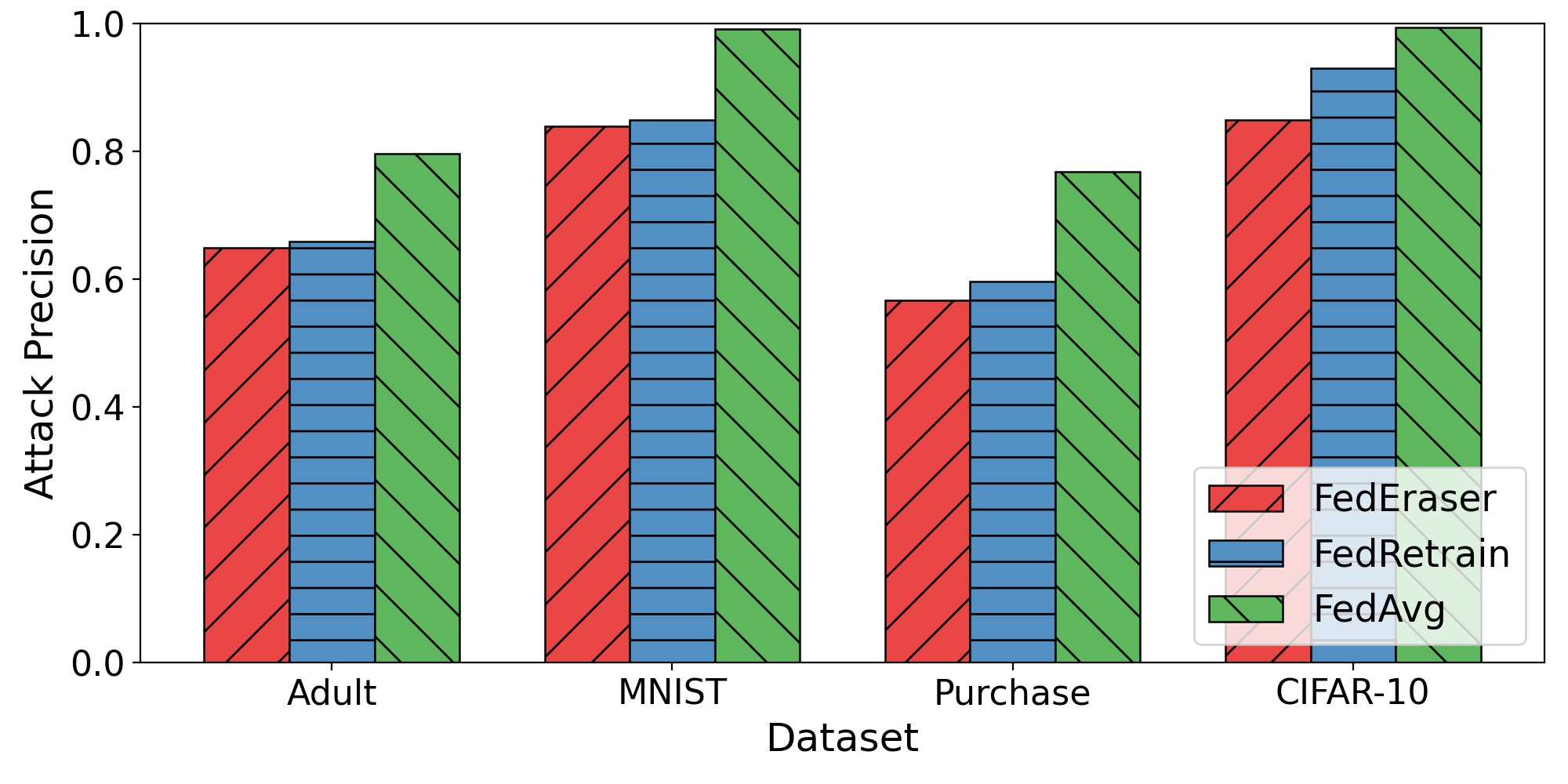}
		\centering
		\label{Figure:mia_rec}
	}

	\caption{Performance of membership inference attacks.}
	\label{Figure:mia_attack}
\end{figure*}

Furthermore, we measure the loss values of the target client's data obtained from different models trained by FedEraser and the comparisons. The experiment results are shown in Table~\ref{Tab:loss}.
In general, the prediction loss of FedEraser is relatively close to that of FedRetrain, and FedAccum has the largest prediction loss among all comparison methods.  For Adult dataset, FedEraser achieves a prediction loss of $5.42 \times 10^{-3}$ that is very close to that of FedAvg and Retrain. However, the loss of FedAccum is $2.7\times$ larger than that of FedEraser. As for MNIST dataset, FedEraser gets a mean loss on the target data of $1.03 \times 10^{-3}$, which is $1.3 \times $ greater than the baseline but $0.8 \times$ smaller than directly accumulating. For Purchase dataset, the loss of FedEraser is $3.85 \times 10^{-3}$, which is much closer to the baseline than that of FedAccum. Besides, FedEraser even achieves a prediction loss of $2.03\times10^{-2}$ that is smaller than that of FedRetrain.

\subsubsection{Evaluation from the Privacy Perspective}\label{Exp.2}

In our experiments, we leverage MIAs towards the target client's data to assess how much information about the these data is still contained in our unlearned model.
Since the attack classifier is trained on the data derived from the original global model, the attack classifier can distinguish the information related to the target data precisely.
\emph{The worse the performance of the MIA is, the less influence of the target data is stored in the global model.}

For executing MIAs towards the unlearned model, we adopt the strategy of shadow model training~\cite{shokri2017membership} to derive the data for construct an attack classifier. For ease of presentation, we treat the original model trained by FedAvg as the shadow model. Then we execute the attack against the global models trained by FedEraser and FedRetrain.

From the results in Fig.~\ref{Figure:mia_attack}, we can see that the attack achieves resemble performances on our unlearned model and the retrained model.
Over all datasets, the inference attacks can only achieve a mean attack precision (resp. recall) around $0.50$ (resp. $0.726$) on the global models reconstructed by FedEraser.

Specifically, for Adult dataset, the attack against the original model can achieve an \emph{F1-score} of $0.714$. The \emph{F1-score} of the attack on the unlearned model (resp. retrained model) is $0.563$ (resp. $0.571$).
As for MNIST and Purchase datasets, it only differs by $0.34\%$ (resp. $0.15\%$) on the \emph{F1-score} difference of the attacks against the unlearned and retrained models.
Besides, compared with FedRetrain, FedEraser can effectively erase the target data even for a complex model trained on CIFAR-10. The inference attack can achieve an \emph{F1-score} of $0.951$ on the original model. Moreover, when attacking against the unlearned model, the \emph{F1-score} can just reach to $0.629$ which is even lower than attack on the retrained model by $2.02\%$.
These results illustrate that the unlearned model's prediction contains little information about the target data just as the retrained model, and FedEraser can remove the influence of the target data from the original global model.

\subsection{Parameter Deviation of Unlearned Model}\label{Exp.3}
In this section, we dive into the global model trained by FedEraser, and analyze the model parameters. To obtain the insight on the parameter deviation between the retrained and the unlearned model, we conduct an experiment by tracking the last layer's weights of these models at each global training epoch. We also compare the parameters of the global model trained by FedAccum.

\begin{figure}[t]
	\centering
	\includegraphics[width=5.5cm]{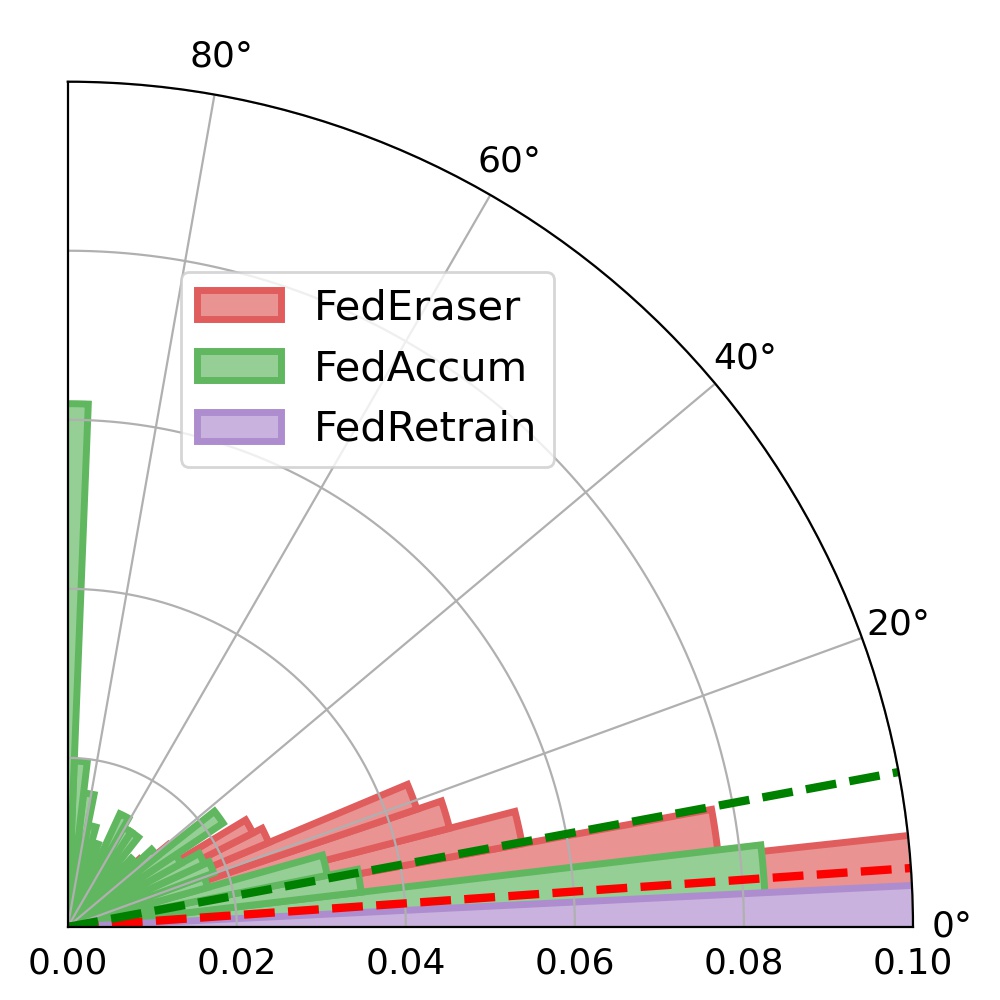}
	\centering
	\caption{Deviation of the global model parameters (Adult).}
	\label{Figure:angle}
\end{figure}

In Fig.~\ref{Figure:angle}, we visualize a histogram of deviation: $\theta = \arccos \frac{{w_{u} \dot w_{r}}}{||w_{u}||||w_{r}||}$, where $w_{u}$ (resp. $w_{r}$) is the last layer weight of the model trained by FedEraser (resp. FedRetrain). The deviation between $w_{a}$ and $w_{r}$ is also represented in Fig.~\ref{Figure:angle}, where $w_{a}$ is the last layer weight of the model trained by FedAccum.

From the results we can observe that the parameters of the unlearned model are much closer to the retrained model than that of the model reconstructed by FedAccum.
As shown in Fig.~\ref{Figure:angle}, the mean angle deviation between $w_{u}$ and $w_{r}$ reaches to $3.99^{\circ}$ (the red dash line in Fig.~\ref{Figure:angle}). The deviation of $w_a$ is higher than that of FedEraser by $2.6\times$ and reaches to $10.56^{\circ}$ (the green dash line in Fig.~\ref{Figure:angle}).
The parameters of our unlearned model are mainly distributed within $30^{\circ}$ difference from that of the retrained model. However, for the accumulated model, there are high deviations ($0^{\circ}\sim90^{\circ}$) between $w_a$ and $w_r$. Furthermore, in the deviation range greater than $30^{\circ}$, it mainly contains the parameters of the accumulated models rather than that of our unlearned models.

\begin{figure*}[t]
	\centering
	\subfigure[Performance (Adult)]{
		\includegraphics[width=5.7cm]{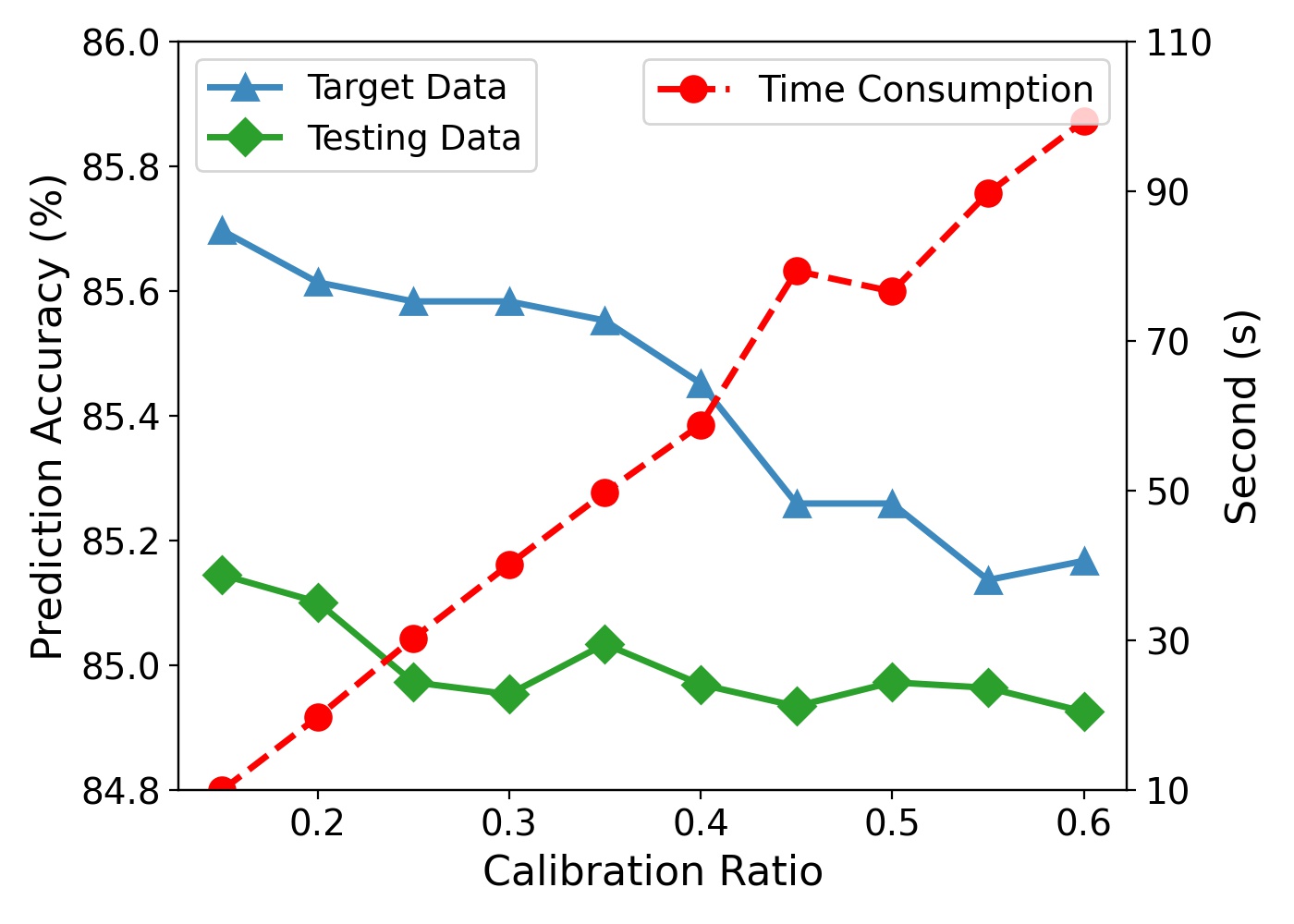}
		\centering
		\label{Figure:cali_ratio_adult}
	}
	\subfigure[Performance (MNIST)]{
		\includegraphics[width=5.7cm]{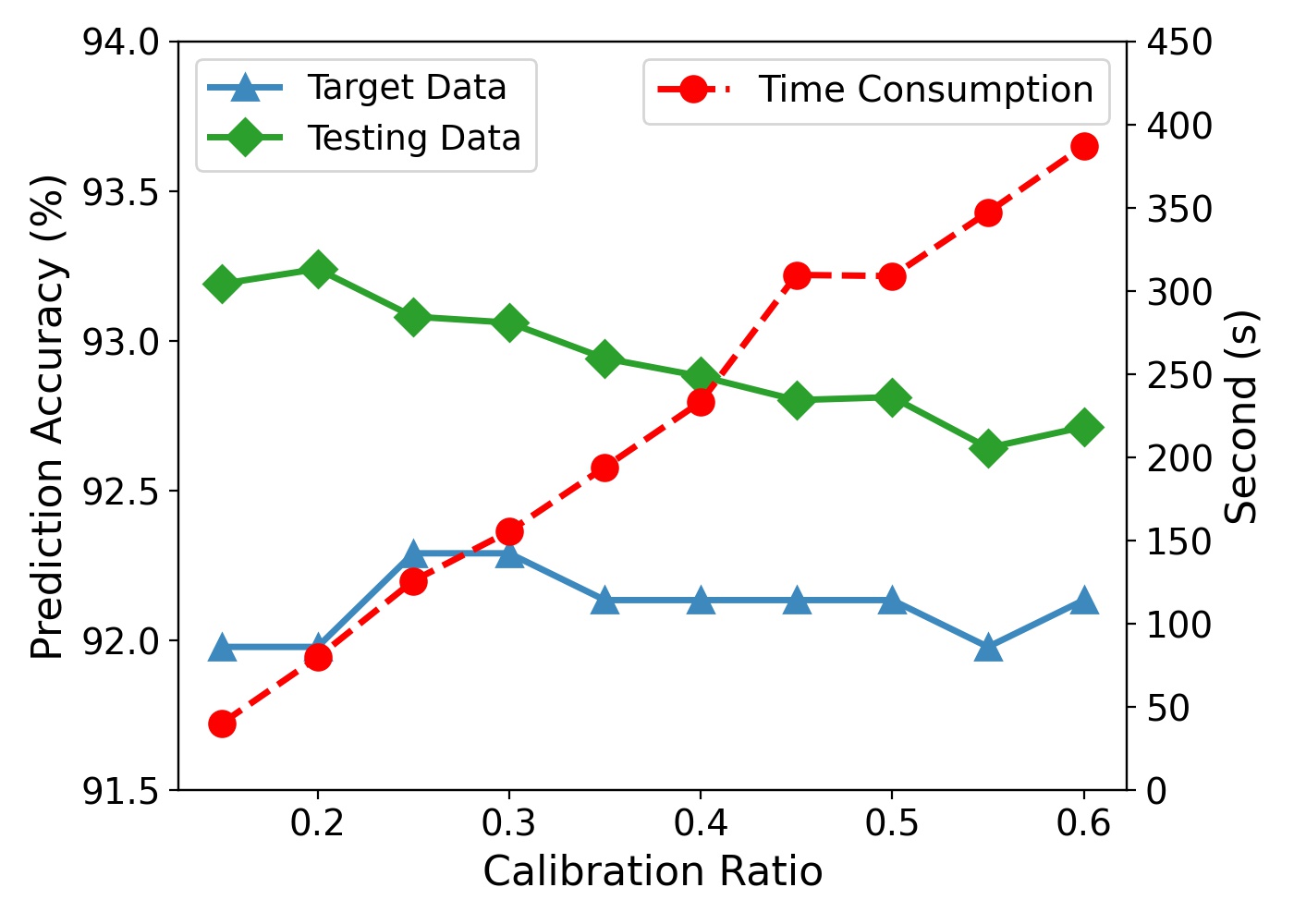}
		\centering
		\label{Figure:cali_ratio_mnist}
	}
	\subfigure[Performance (Purchase)]{
		\includegraphics[width=5.7cm]{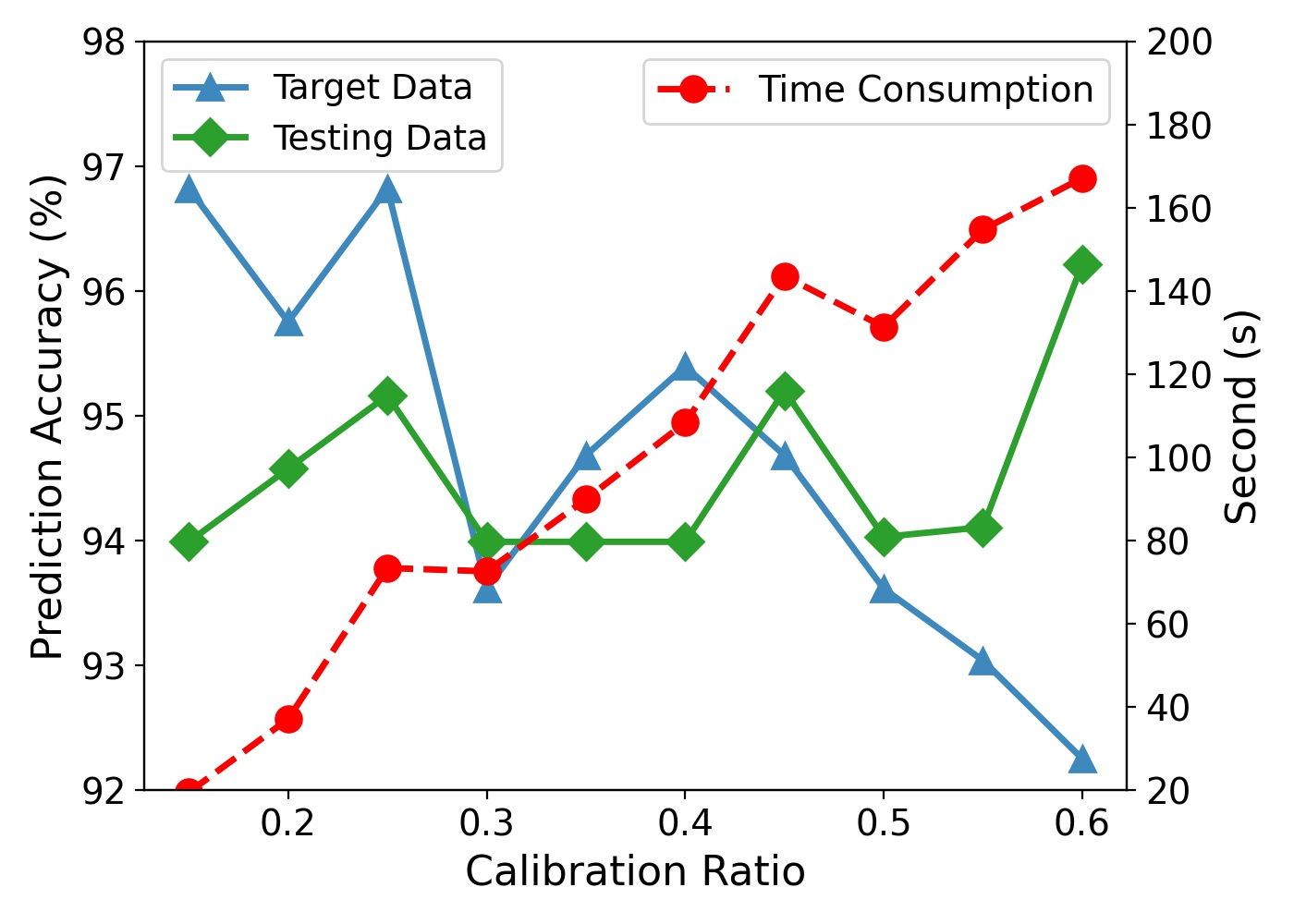}
		\centering
		\label{Figure:cali_ratio_purchase}
	}
	\caption{The impact of calibration ratio on performance of FedEraser.}
	\label{Figure:cali_ratio}\vspace{5pt}
\end{figure*}

\begin{figure*}[t]
	\centering
	\subfigure[Performance (Adult)]{
		\includegraphics[width=5.7cm]{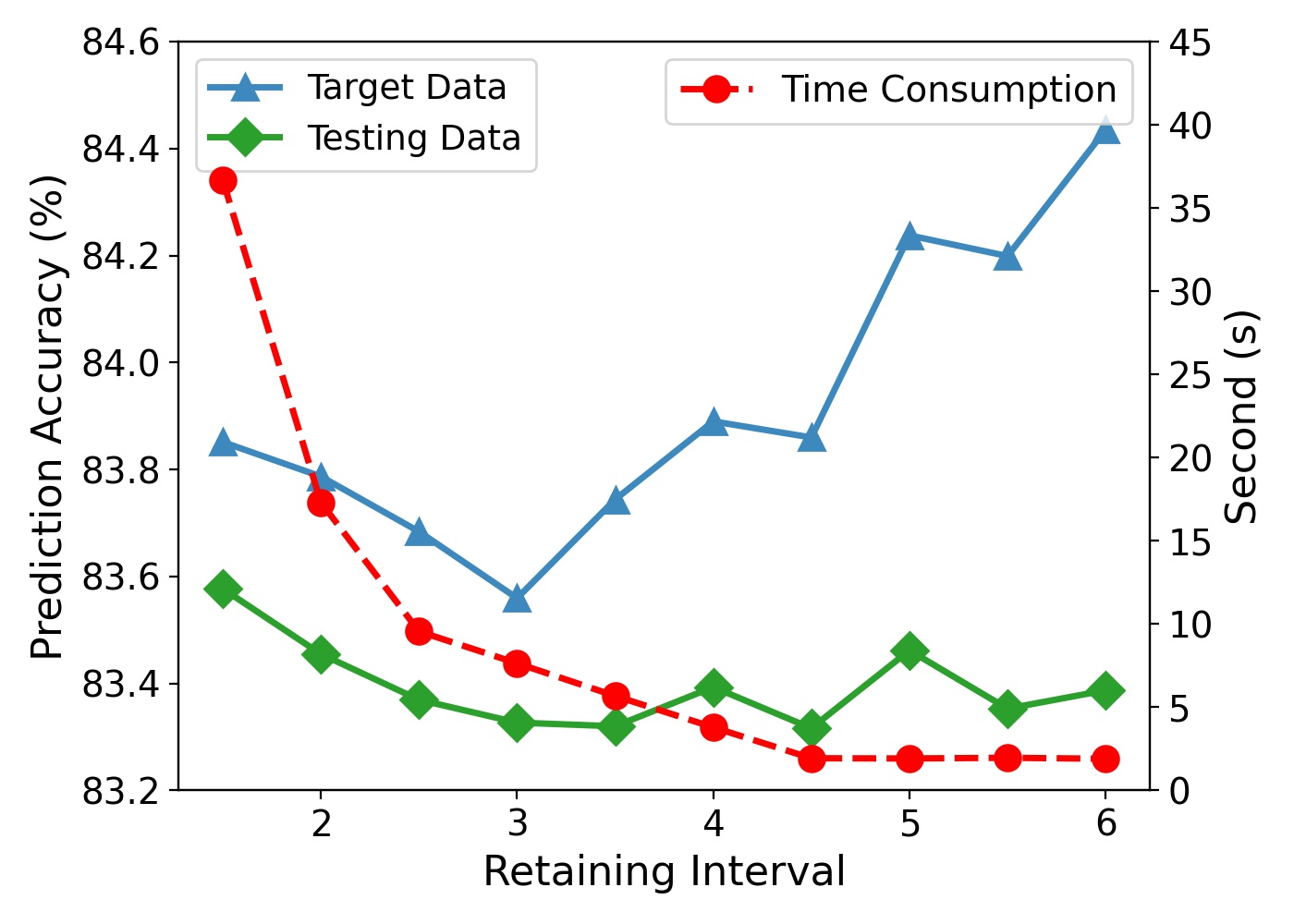}
		\centering
		\label{Figure:cali_int_adult}
	}
	\subfigure[Performance (MNIST)]{
		\includegraphics[width=5.7cm]{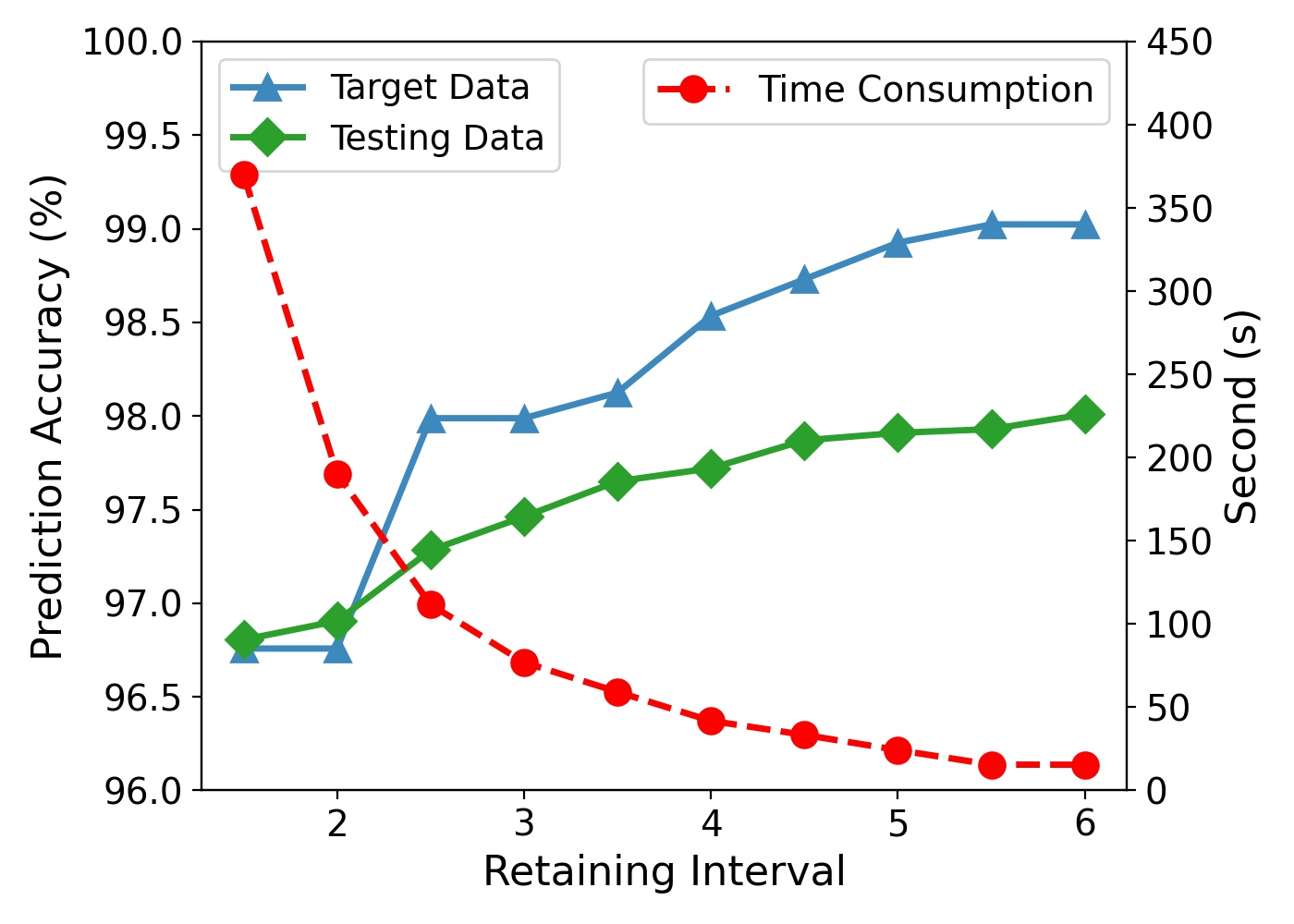}
		\centering
		\label{Figure:cali_int_mnist}
	}
	\subfigure[Performance (Purchase)]{
		\includegraphics[width=5.7cm]{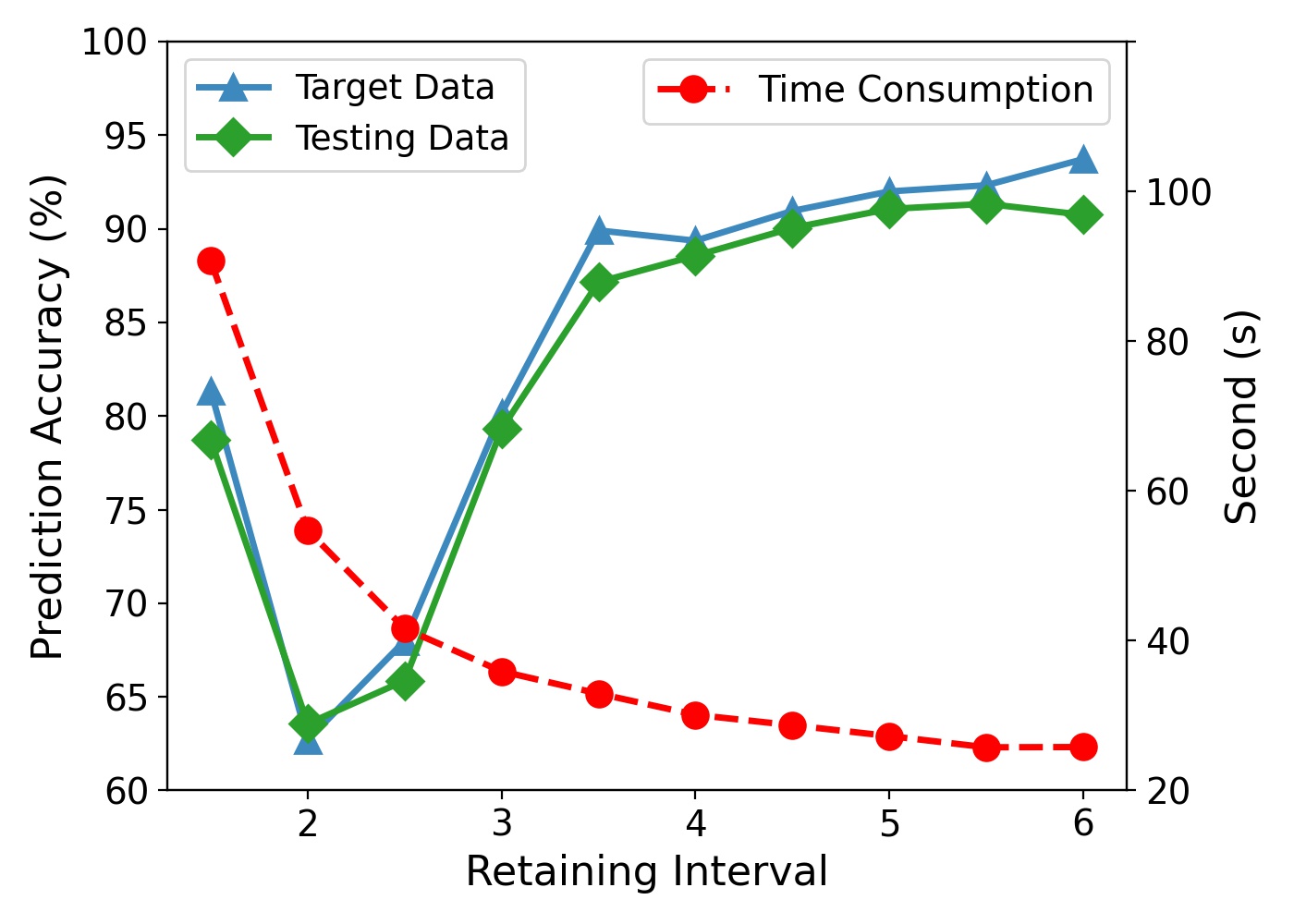}
		\centering
		\label{Figure:cali_int_purchase}
	}
	\caption{The impact of retaining interval on performance of FedEraser.}
	\label{Figure:cali_int}
\end{figure*}

\subsection{Impact of the Local Calibration Ratio}\label{Exp.4}
To quantify the impact that the local calibration ratio $\verb"r"$ has on the performance of FedEraser, we reconstruct the original global model trained on three different datasets with $\verb"r"$ from $0.1$ to $1$. When $\verb"r"=1.0$, FedEraser will degenerate to FedRetrain, which is the baseline in our experiments.

Fig.~\ref{Figure:cali_ratio} shows the relationship between the calibration ration and the performance of FedEraser. The relationship is a little complex, but in general, as $\verb"r"$ increases, the prediction accuracy on the target data becomes worse but the reconstruction time increases almost linearly.
Specifically, for Adult dataset, when $\verb"r"=0.1$, FedEraser achieves a prediction accuracy of $85.8\%$ (resp. $84.9\%$) on the target (resp. testing) data, and the time consumption for unlearned model reconstruction is just $10.1 s$, which attains a speed-up  of $10\times$ compared with FedRetrain. When $\verb"r"$ increases to $1.0$, the accuracy on the target (resp. testing) data degrades by $0.5\%$ (resp. $0.2\%$). However, the reconstruction time increases and reaches to $100.1 s$.

For MNIST dataset, FedEraser can achieve a prediction accuracy of $91.9\%$ (resp. $93.5\%$) on the target (resp. testing) data when $\verb"r"=0.1$. In this case, the time consumption of FedEraser is merely $40.3 s$ which is faster than FedRetrain by $9.6 \times$. As $\verb"r"$ increases, the prediction accuracy on the testing data decreases slightly, which is not the case for that on the target data. Specifically, with different calibration ratio, the prediction accuracy on the target data fluctuates around $92.1\%$ with a standard deviation of $0.0011$.

As for Purchase dataset, when treating the target data, FedEraser confronts a reduction of $5.2\%$ in the prediction accuracy with the increasing calibration ratio. Nevertheless, in this case, the testing accuracy of FedEraser increases from $93.9\%$ to $96.2\%$. As for the reconstruction time of the unlearned model, FedEraser only needs $19.6 s$ when $\verb"r"=0.1$, yielding a speed-up of $8.4\times$ compared with the time consumption of FedRetrain.


\subsection{Impact of the Retaining Interval}\label{Exp.5}
In this section, we evaluate the performance of FedEraser on three different datasets with retaining interval $\Delta t$ increasing from $1$ to $10$. The relationship between the performance of FedEraser and the retaining interval is demonstrated in Fig.~\ref{Figure:cali_int}. From the results we can find that with the increasing retaining interval, the time consumption of FedEraser decays while the prediction accuracy on the target data improves better and better. One possible reason for this phenomenon is that with a large retaining interval, a part of the influences of the target data are still remained in the unlearned model.

Recalling that the objective of FedEraser is to eliminate the influences of a certain client's data in the original global model. These influences are involved by training the original model on these data and could help this model accurately classify the target data.
Therefore, the higher accuracy on the target data, the worse performance of FedEraser achieves.
According to the results in Fig.~\ref{Figure:cali_int}, FedEraser brings about a poor performance but a obvious  when the retaining interval is set to a large number.

\begin{figure*}[t]
	\centering
	\subfigure[Prediction Accuracy (Purchase)]{
		\includegraphics[width=7.7cm]{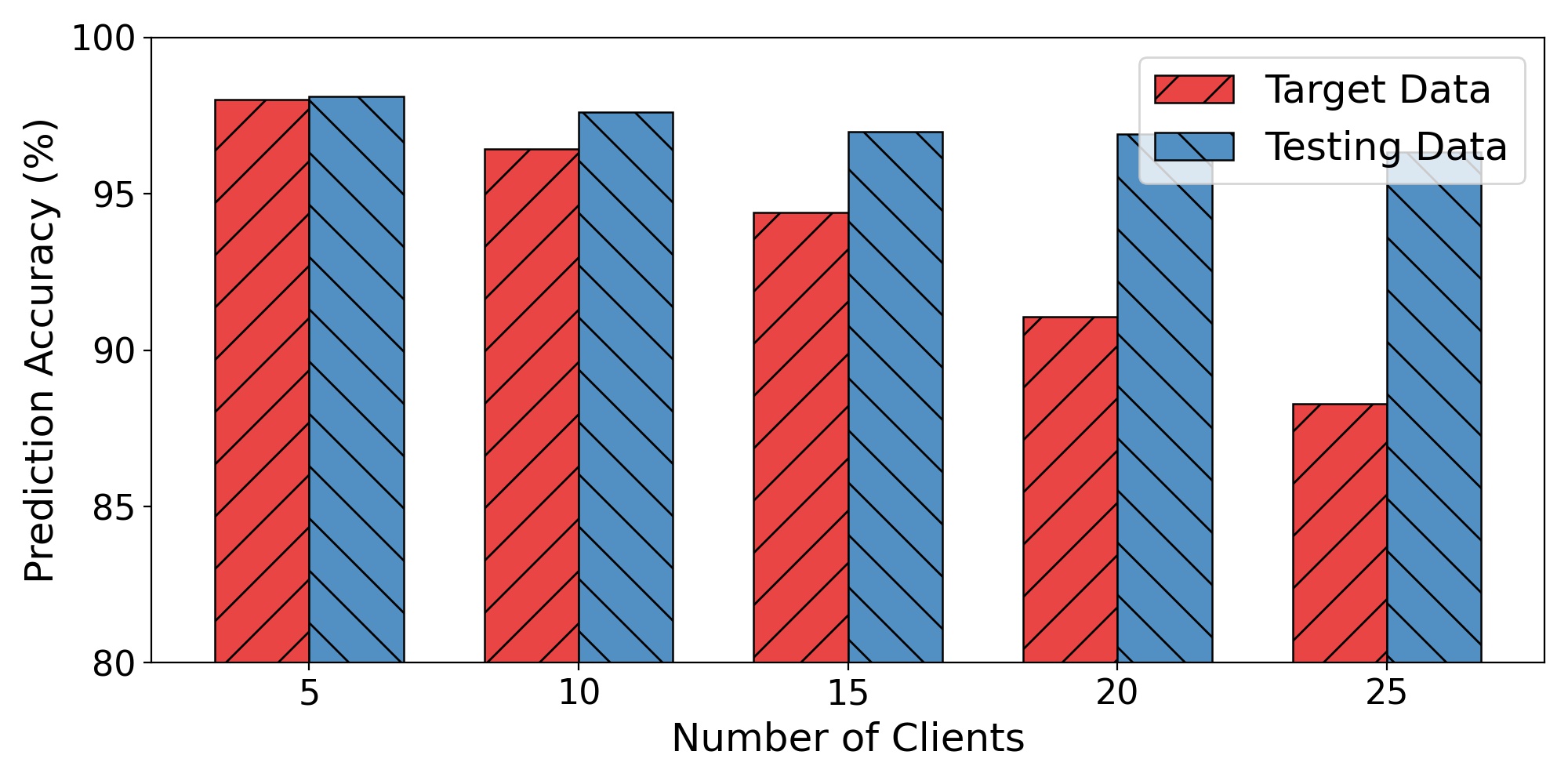}
		\centering
		\label{Figure:num_client_purchase}
	}
	\subfigure[Prediction Accuracy (CIFAR-10)]{
		\includegraphics[width=7.7cm]{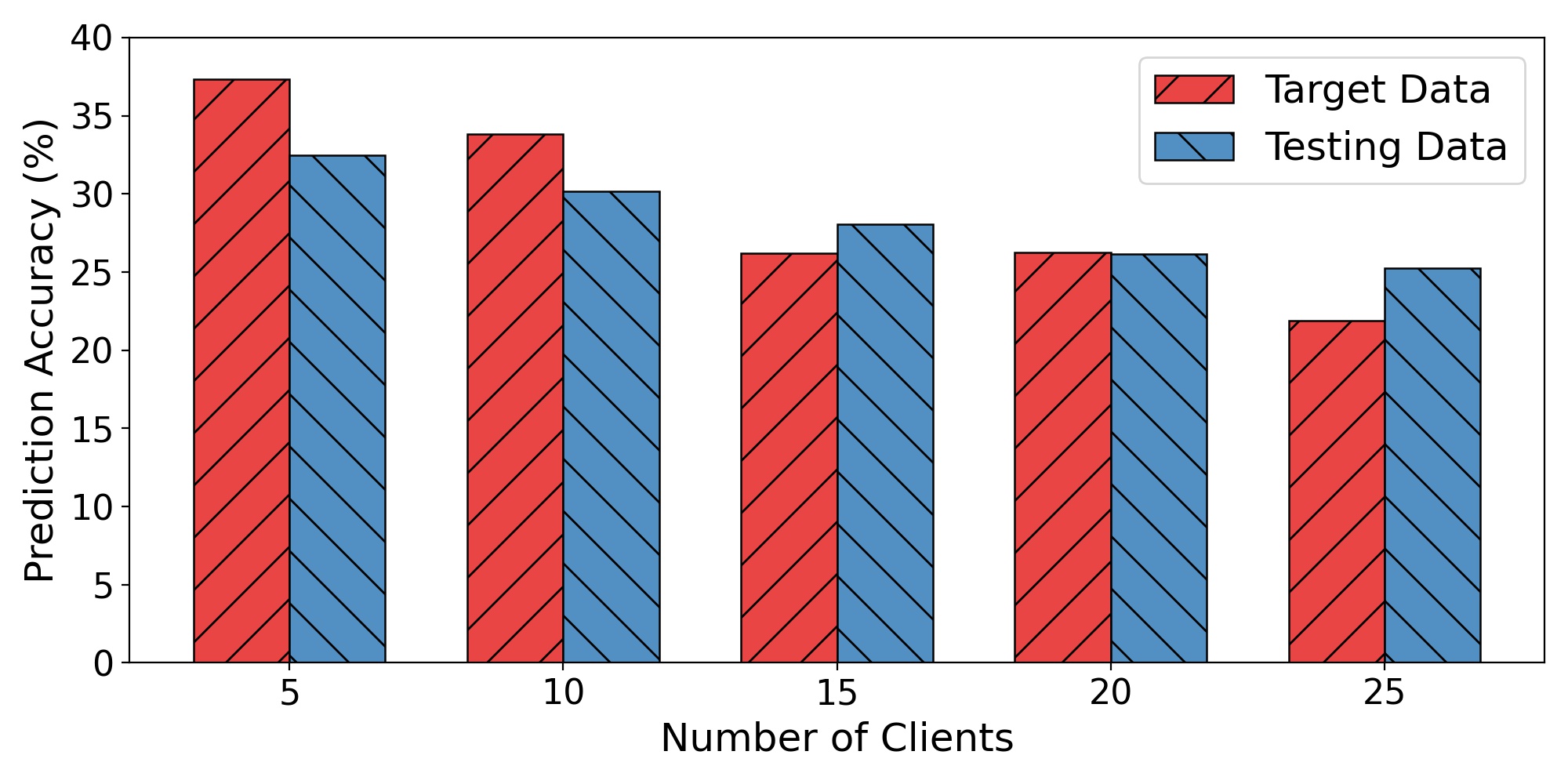}
		\centering
		\label{Figure:num_client_cifar}
	}
	
	\caption{The impact of the number of federated clients.}
	\label{Figure:num_client}
\end{figure*}

As shown in Fig.~\ref{Figure:cali_int_adult}, with the retaining interval increasing, the prediction accuracy on the target data increases from $83.8\%$ to $84.4\%$ but the testing accuracy decreases by $0.21\%$. When $\Delta t=1$, FedEraser spends $36.7 s$ to reconstruct the original global model. But when $\Delta t=10$, it consumes  $19.1 s $ to derive the unlearned model which brings a $12\times$ speed-up.

As shown in Figs.~\ref{Figure:cali_int_mnist} and~\ref{Figure:cali_int_purchase}, the prediction accuracy of our unlearned models prompts both on the target and testing data when FedEraser confronts MNIST and Purchase datasets. For MNIST dataset, our unlearned model can achieve a prediction accuracy of $96.7\%$ (resp. $96.8\%$) on the target (resp. testing) data. As the interval increases, the accuracy on the target data increases by $2.3\%$, while that on the testing data increases by $1.2\%$.
As for Purchase dataset, the accuracy of our unlearned model increases from $81.3\%$ to $93.7\%$ on the target data, and the accuracy also grows from $80.1\%$ to $92.1\%$ on the testing data.
As for the time consumption, FedEraser can yield $12 \times$ speed-up on both datasets when $\Delta t=10$.

\subsection{Impact of the Number of Federated Clients}\label{Exp.6}
In this section, we evaluate the performance of FedEraser on the Purchase and CIFAR-10 datasets with different number of federated clients. From the results in Fig.~\ref{Figure:num_client}, we can observe that the performance of the unlearned model gradually degrades with the increasing number of clients.
Specifically, for Purchase dataset (c.f. Fig.~\ref{Figure:num_client_purchase}), when there are $5$ federated clients, the model reconstructed by FedEraser can achieve a prediction accuracy of $98.0\%$ (resp. $98.1\%$) on the target (resp. testing) data. When the number of clients increasing to $25$, the prediction accuracy decreases by $1.8\%$ but can still reach $96.3\%$. However, for the target data, the performance of FedEraser would degrade by $9.7\%$ and achieve a prediction accuracy of $88.3\%$.

For CIFAR-10 dataset (c.f. Fig.~\ref{Figure:num_client_cifar}), with total $5$ federated clients, the unlearned model can achieve a prediction accuracy of $32.4\%$ (resp. $27.3\%$) on the testing (resp. target) data. As the number of clients increasing, our unlearned model performs worse on both the testing and target data gradually. When there are $25$ clients,
the prediction accuracy on the target (resp. testing) data decreases by $5.4\%$ (resp. $7.2\%$).

Overall, all the results demonstrate that FedEraser can achieve a satisfied performance in different datasets with different settings. In general, if a data sample has taken part in a model's training process, it would leave its unique influence on this model so that the model can correctly classify on it. Therefore, the prediction accuracy on the target data can measure how much influence of these data left in the unlearned model.
The less influence of the target data leaves on the model reconstructed by FedEraser, the lower prediction accuracy on the target data this model can achieve, and the better performance of FedEraser will be.


\section{Related Work}\label{Sec:RelatedWork}

\subsection{Machine Unlearning}

The term ``machine unlearning'' is introduced by Cao et al.~\cite{cao2015spmul},
where an efficient forgetting algorithm in the restricted context of statistical query learning is proposed.
Thereafter, machine unlearning for different ML models have been explored.
Ginart et al.~\cite{ginart2019making} examine the problem of data removal algorithm for stochastic algorithms, in particular for variants of $k$-means clustering, but cannot be applied to supervised learning.
Izzo et al.~\cite{izzo2020approximate} focus on supervised linear regression and develop the projective residual update technique that scales linearly in the dimension of the data.
Baumhauer et al~\cite{baumhauer2020machine} propose a forgetting procedure for logit-based classification models by applying linear transformation to the output logits, but do not remove information from the weights.
Most recently, Bourtoule et al.~\cite{bourtoule2020spmul} introduce a more general algorithm named SISA, which takes advantage of sharding and slicing during the training.
Nevertheless, existing machine unlearning studies focus on ML models in traditional centralized settings, and is ineligible for unlearning in FL scenarios.



\subsection{Differential Privacy}
Differential privacy~\cite{RN9} provides a way to preserve the privacy of a single sample in a dataset such that an upper bound on the amount of information about any particular sample can be obtained.
There have been a series of differentially private versions of ML algorithms, including linear models~\cite{Yan2017Differentially}, principal component analysis~\cite{mechanism2016aaai}, matrix factorization~\cite{imtiaz2018distributed}, and DNN~\cite{abadi2016deep}, the parameters of which are learned via adding noise in the training phase.
In the setting of data forgetting, however, the removal is expected to be done after the training.

Drawing on the indistinguishability of differential privacy, Guo et al.~\cite{guo2020certified} define the notion of $\epsilon$-certified removal and provide an algorithm for linear and logistic regression.
Golatkar et al.~\cite{golaktar2020eternal} propose a selective forgetting procedure
for DNNs by changing information (adding noises) in the trained weights.
They further extend this framework to disturb activations~\cite{golatkar2020forgetting}, using
a neural tangent kernel based scrubbing procedure. 
The major challenge in differential privacy based unlearning is how to balance the protected information and the model utility.

\subsection{MIAs against ML Models}
MIA against ML models was first studied by Shokri et al.~\cite{shokri2017membership}, where multiple shadow models with the same structure as the victim model are constructed to facilitate training the attack models. Later on, Salem et al.~\cite{salem2018ml} show that it is possible to achieve the resemble attack performance with only one shadow model. Liu et al.~\cite{liu2019socinf} leverage the idea of generative adversarial networks (GAN) to train a mimic model instead of the shadow model.
Apart from mimicking the prediction behavior, other knowledge of the victim model is adopted to launch MIAs, including the training loss~\cite{yeom2018overfit}, model parameters~\cite{melis2019exploiting}, model gradients~\cite{nasr2019comprehensive}, and output distributions~\cite{hui2021practical}.
Intuitively and universally acknowledged, such attacks can serve as one of the best befitting manners for measuring the quality of unlearning~\cite{sommer2020towards,sablayrolles2020radioactive}, especially given few eligible metrics for evaluating the performance of such attacks.
Therefore in this paper, we also adopt MIAs to evaluate the effectiveness of FedEraser.


\section{Conclusion}\label{Sec:Conclusion}
In this paper, we have presented FedEraser, the first federated unlearning methodology that can eliminate the influences of a federated client's data on the global model while significantly reducing the time consumption used for constructing the unlearned model.
FedEraser is non-intrusive and can serve as an opt-in component inside existing FL systems.
It does not involve any information about the target client, enabling the unlearning process performed unwittingly.
Experiments on four realistic datasets demonstrate the effectiveness of FedEraser, with an obvious speed-up of unlearning compared with retraining from scratch.
We envision our work as an early step in FL towards compliance with 
legal and ethical criteria in a fair and transparent manner.
There are abundant interesting directions opened up ahead, e.g., instance-level federated unlearning, federated unlearning without client training, federated unlearning verification, to name a few.
We plan to investigate these appealing subjects in the near future.

\section*{Acknowledgement}
This work was supported in part by the National Natural Science Foundation of China under Grants 61872416 and 62002104; by the Fundamental Research Funds for the Central Universities of China under Grant 2019kfyXJJS017; by the Natural Science Foundation of Hubei Province of China under Grant 2019CFB191; and by the special fund for Wuhan Yellow Crane Talents (Excellent Young Scholar). The corresponding author of this paper is Chen Wang.

\bibliographystyle{IEEEtran}
\bibliography{mybib}

\end{document}